# High-Load-Density Electro-Permanent Magnetic Foot with Controllable Adhesion for Quadruped Wall-Climbing Robots

An Li, Bo Tao, I-Ming Chen and Han Ding

*Abstract*— **To enable reliable climbing locomotion of quadruped robots on ferromagnetic surfaces, this paper presents a high-load-density electro-permanent magnetic foot with controllable adhesion, featuring force-feedback circular Halbach-net electro-permanent magnet (CHN-EPM) adhesion units and a magnetization control system. Due to its three-dimensional magnetic circuit structure and flux-concentration effect, the CHN-EPM enables a distributed parallel magnetic flux path with enhanced flux utilization, resulting in reduced sensitivity to air-gap variations and allowing effective adhesion to be maintained even under partial contact conditions. The proposed CHN-EPM generates a maximum adhesion force exceeding 1000 N with a load-to-weight ratio over 200:1. A magnetization driver and a two-stage pulse current control strategy are developed to regulate the excitation current amplitude and duration, enabling accurate and reliable magnetization. By incorporating a flexible pressure sensor for contact force feedback, the system can effectively monitor attachment and detachment states, ensuring robust adhesion switching under uncertain contact conditions. The proposed system is integrated into a commercial quadruped robot (Unitree GO2), demonstrating high-load adhesion on ceiling and vertical-wall surfaces and stable locomotion on painted, perforated, and curved ferromagnetic surfaces.**

## I. Introduction

LARGE and complex steel structures, such as wind turbine towers, offshore oil platforms, and ship hulls, are critical in many industrial applications. Wall-climbing robots provide an effective solution for mobile operations on these surfaces, where conventional equipment is limited by scale and accessibility [1]. Various adhesion and attachment strategies have been explored for wall-climbing robots, including vacuum negative pressure adhesion [2][3], bioinspired spiny attachment [4], electroadhesion [5], electromagnetic adhesion [6], and magnetic adhesion [7]–[9]. Among these approaches, magnetic adhesion is particularly suitable for ferromagnetic environments due to its high load capacity, low energy consumption, and flexible configuration.

In terms of locomotion, magnetic adhesion wall-climbing robots can be broadly categorized into continuous systems (e.g., wheeled and tracked) and discrete systems (e.g., inchworm and legged robots) [10]-[16]. Compared to continuous mechanisms, legged configurations offer superior adaptability to complex geometries, including obstacles, curved surfaces, and multi-plane transitions. However, their performance critically depends on the reliable attachment and detachment of magnetic feet during locomotion.

Existing magnetic adhesion solutions exhibit notable limitations. Permanent magnets require additional mechanical mechanisms for detachment, increasing system complexity and response time [17]. Electromagnets enable fast switching but suffer from large size, high power consumption, and thermal issues under continuous current excitation [6]. Recently, electro-permanent magnets (EPMs) have emerged as a promising alternative, combining high load density with near-zero power consumption by using pulse currents to switch magnetization states [18]-[20].

Nevertheless, two key challenges remain for current EPM-based magnetic feet. The first is to improve adhesion force density through an optimized electromagnetic topology, which is critical for achieving high load capacity in compact designs. Prior designs, such as parallel EPM (P-EPM) [21] and square EPM (S-EPM) [22], have improved switching efficiency by reducing coil length. However, they still exhibit limited adhesion force density and high sensitivity to air-gap variations, ferromagnetic wall thickness and partial contact conditions, leading to degraded performance under imperfect contact conditions. The second, and more critical challenge, lies in achieving controllable and reliable magnetization under real-world conditions. In practical applications, the excitation coils in EPM units are often exposed to mechanical impact, making them susceptible to damage, which may lead to loss of magnetization controllability or even complete failure of the adhesion system. Besides, current EPM units are typically driven by directly connecting to a power source through switching circuits, where the magnetization and demagnetization processes rely on uncontrolled pulse currents [23]. As a result, both the amplitude and duration of the pulse

This manuscript is a preprint version prepared for early dissemination. This work was supported in part by the National Natural Science Foundation of China (NSFC) under Grant 52305022 and Grant 62293514.
(Corresponding author: Bo Tao, e-mail: taobo@hust.edu.cn).

An Li and I-Ming Chen are with the School of Mechanical and Aerospace Engineering, Nanyang Technological University, Singapore.

Bo Tao and Han Ding are with the School of Mechanical Science and Engineering, Huazhong University of Science and Technology, Wuhan430074, China.

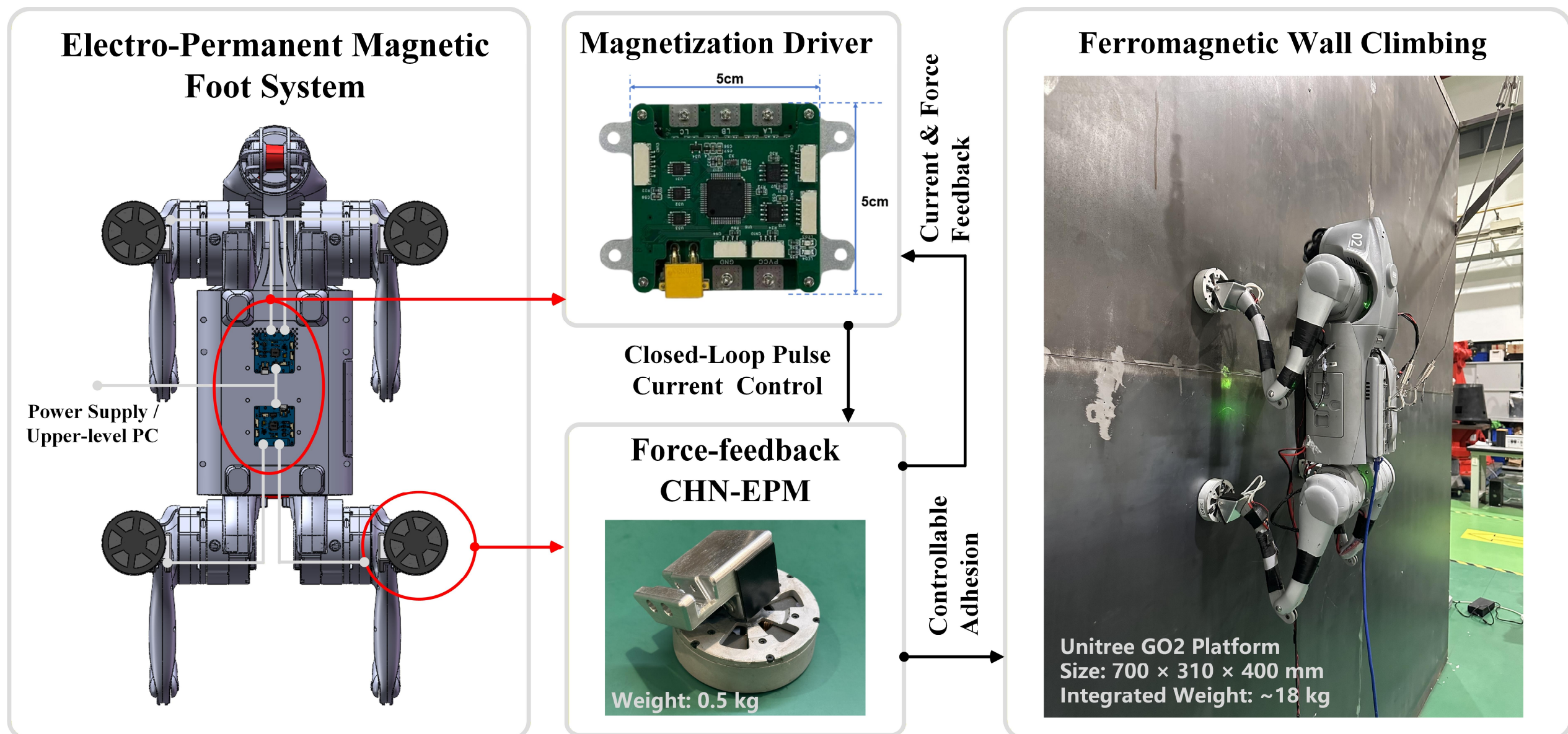

**Fig. 1.** Overview of the electro-permanent magnetic foot control system in a quadruped wall-climbing robot.

current cannot be precisely regulated. This leads to two major issues: insufficient or excessive excitation may cause magnetization failure or instability, while uncontrolled pulses introduce significant energy loss, which becomes critical for repeated switching during long-duration operations.

In addition to electrical controllability, magnetization is highly sensitive to contact conditions, as air gaps or insufficient contact hinder effective magnetic field establishment and reduce adhesion reliability. To mitigate the challenge of contact uncertainty, several sensing-based approaches have been explored. For example, torque and angle feedback have been employed for closed-loop contact detection in a six-limbed inspection robot equipped with electro-permanent magnets [24]. Other studies have explored various sensing approaches, including crack-based tactile sensors for contact and ground reaction sensing in legged robots [25] , triboelectric tactile sensors for monitoring magnetic adhesion states in wall-climbing robots [26], membrane pressure sensors for measuring suction-cup adsorption force [27], and self-inductive sensing claws for electromagnetically adhering climbing robots [28]. However, these methods are either designed for non-magnetic adhesion systems or are not directly integrated into the magnetization control loop of electro-permanent magnetic feet.

Therefore, this paper presents a high-load-density electro-permanent magnetic foot with controllable adhesion for quadruped wall-climbing robots. A circular Halbach-net electro-permanent magnet (CHN-EPM) topology is proposed [29], which leverages a three-dimensional magnetic circuit and flux-concentration effect to enable a distributed parallel magnetic flux path with enhanced flux utilization. This configuration reduces the sensitivity of air-gap flux density to gap variations, thereby maintaining effective adhesion under partial contact conditions. The proposed design integrates force-feedback CHN-EPM adhesion units with a magnetization control system to achieve both enhanced adhesion performance and reliable magnetization control.

## II. Design and Principle of the Magnetic Foot

In this section, the design and operating principle of the proposed magnetic foot are presented. The magnetic foot system consists of four force-feedback CHN-EPM units and two magnetization drivers. The detailed structure, operating principle, and performance evaluation are presented in the following subsections.

### *A. System Architecture of the Magnetic Foot*

As illustrated in **Fig. 1** , the proposed magnetic foot control system consists of force-feedback CHN-EPM units and magnetization drivers, which together enable controllable magnetic adhesion for quadruped wall-climbing robots. The force-feedback CHN-EPM units are mounted at the feet of the robot and directly interact with the wall surface, generating magnetic adhesion force for attachment and detachment during locomotion. Meanwhile, the magnetization drivers are responsible for generating pulse current signals to control the magnetization and demagnetization of the CHN-EPM units, thereby regulating the adhesion force at the foot–wall interface.

The force-feedback CHN-EPM units also provide real-time measurements of the contact force through embedded pressure sensors, which are fed back to the magnetization drivers for adhesion state monitoring and safety assurance. Moreover, the magnetization drivers implement closed-loop regulation of the pulse current, including both amplitude and duration, based on real-time current feedback to ensure reliable magnetization under varying contact conditions.

In the system implementation, the magnetization drivers are installed within the robot body (e.g., the trunk of the Unitree GO2 platform) and communicate with the upper-level controller via CAN bus. The drivers receive adhesion control commands from the controller and transmit the force feedback signals for system monitoring. Powered by an external supply battery, each magnetization driver is connected to two magnetic feet through dedicated pulse power and sensing interfaces routed along the robot body. This configuration enables coordinated control of multiple magnetic feet while maintaining a compact and scalable system architecture.

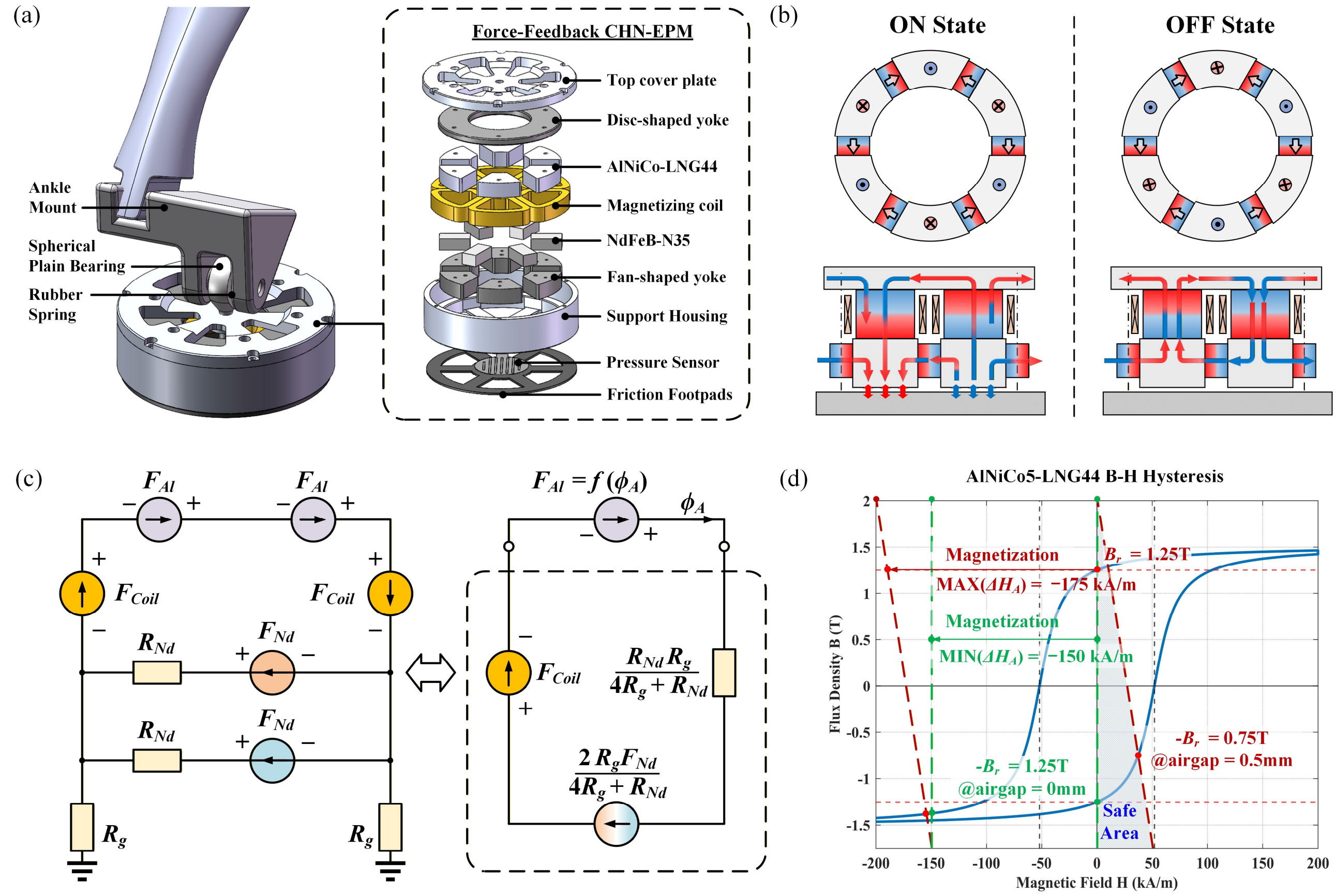

**Fig. 2.** Structure and magnetization control principle of the CHN-EPM: (a) structural configuration of the magnetic foot; (b) magnetic flux distribution in ON/OFF states; (c) equivalent magnetic circuit model; (d) nonlinear B–H characteristics and magnetization condition.

*B. Magnetization Control Principle of the CHN-EPM*

As shown in **Fig. 2** (a), the structural configuration of the magnetic foot is introduced. The magnetic foot mainly consists of a CHN-EPM unit, a passive ankle mechanism, and a force-feedback footpad module. The CHN-EPM is composed of a disc-shaped silicon steel yoke, multiple fan-shaped yokes, NdFeB permanent magnets, AlNiCo semi-hard magnets, and excitation coils. It is connected to the robot leg via a passive ankle mechanism incorporating spherical joints and extension springs, providing compliance and sufficient degrees of freedom for surface adaptation. A flexible pressure sensor is mounted beneath the support housing for force feedback. In addition, a perforated silicone rubber layer, with the same thickness as the fan-shaped yoke bottom, is attached to enhance friction performance.

The controllable adhesion of the proposed CHN-EPM is achieved by regulating the magnetization direction of the AlNiCo magnets relative to the NdFeB magnets. As illustrated in **Fig. 2**(b), when the magnetization polarity of AlNiCo is aligned with that of NdFeB, the magnetic flux is reinforced and guided through the fan-shaped silicon steel yokes into the ferromagnetic surface, producing a strong adhesion force (ON state). Conversely, when the polarity of the AlNiCo magnets is reversed, most of the magnetic flux is confined within the magnetic circuit, significantly reducing the external magnetic field, and releasing the adhesion force (OFF state). Owing to the circular topology, multiple fan-shaped yokes form parallel magnetic flux paths, which improve robustness under partial contact conditions.

To quantitatively analyze the magnetization behavior, an equivalent magnetic circuit model is established, as shown in **Fig. 2** (c). Owing to the symmetry of the CHN-EPM, adjacent excitation coils carry equal and opposite currents, forming a closed magnetic flux loop composed of AlNiCo magnets, NdFeB magnets, silicon steel yokes, and the air gap. Accordingly, the air-gap flux can be expressed as a function of the magnetomotive forces (MMFs) of the AlNiCo and NdFeB magnets,

$$\phi_g = \frac{2F_{Nd}R_{Al} + F_{Al}R_{Nd}}{4R_gR_{Al} + R_gR_{Nd} + R_{Al}R_{Nd}} \tag{1}$$

where $F_{\mathrm{Al}}$ and $F_{\mathrm{Nd}}$ denote the MMFs of the AlNiCo and NdFeB magnets, respectively. $R_{\mathrm{Al}}$ and $R_{\mathrm{Nd}}$ are their equivalent reluctances, and $R_g$ is the equivalent air-gap reluctance between the CHN-EPM and the ferromagnetic surface.

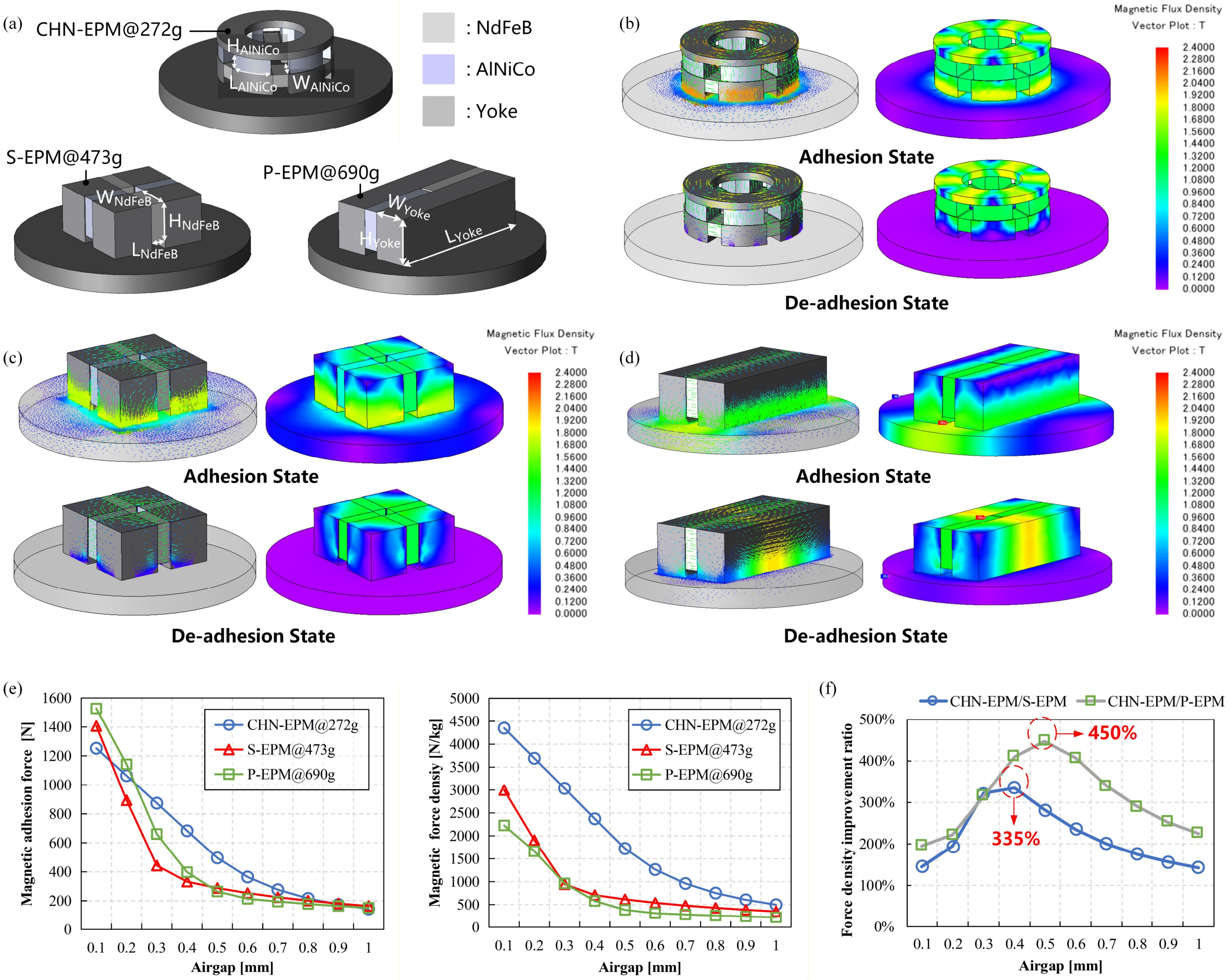

**Fig. 3.** Performance comparison of CHN-EPM, S-EPM, and P-EPM: (a) structural configurations and mass comparison; (b)–(d) magnetic flux distributions under adhesion and de-adhesion states; (e) adhesion force and force density comparison versus air gap and (f) improvement ratio of adhesion force density versus air gap.

Under the de-adhesion state ($\phi_g = 0$), to ensure the CHN-EPM exhibits no magnetism to the wall, it can be derived that $S_{Al} = 2S_{Nd}$, where $S_{Al}$ and $S_{Nd}$ represent the pole area of the AlNiCo, NdFeB magnets.

Under the adhesion state, assuming the relative magnetic permeability $\mu_{r,Nd}$ of NdFeB is constant and $\mu_{r,Al}$ are the relative magnetic permeability of AlNiCo, the magnetic flux density of the air gap can also be derived as,

$$B_g = \frac{\phi_g}{S_g} = \frac{B_{r,Nd} + B_{r,Al}}{\frac{S_g}{S_{Al}} + \frac{2\mu_{r,Al} l_g}{l_{Al}} + \frac{l_g}{l_{Nd}}} \tag{2}$$

where $B_{r,\,Al}$ and $B_{r,\,Nd}$, the magnets remanence of AlNiCo and NdFeB magnet. $S_g$ is the pole area of the airgap. $l_{Al}$, $l_{Nd}$ and $l_g$ represent their respective lengths.

For a CHN-EPM with $n$ fan-shaped yokes, the holding force can be derived using the Maxwell tensor,

$$F_{mag} = \frac{nB_g^2 S_g}{2\mu_0} \tag{3}$$

The above equations indicate that under the adhesion state, the adhesion force is directly related to the magnetic flux density in the air gap and is highly sensitive to the air-gap length. As the air gap increases, both the magnetic flux density and the adhesion force decrease.

The magnetization process requires a pulse current through the excitation coils to reverse the polarity of the AlNiCo magnets. To analyze the required current under varying conditions, the nonlinear B–H characteristics of the AlNiCo magnets are considered, as illustrated in **Fig. 2** (d). The magnetization field is defined as a nonlinear function of the magnetic flux density, i.e., $B_A = f(H_A)$. Based on the equivalent port network model, the magnetizing field of the AlNiCo magnets can be expressed as,

$$H_A = \frac{2k_2 / \mu_0}{4 + k_1}(B_{r,Nd} - B_A) - \frac{NI_c}{l_A} \tag{4}$$

where $k_1 = l_N / l_g$, $k_2 = l_N / l_A$, $l_N$ and $l_A$ are the length of the NdFeB and AlNiCo magnets, $l_g$ is the air gap width, $B_{r,Nd}$ is the residual induction of NdFeB, $I_c$ is the coil current, and $N$ is the number of coil turns.

By varying the air-gap length, the relationship between the magnetization current and the magnetic state can be obtained. Under ideal contact conditions (i.e., negligible air gap), the required magnetization current is minimized and can be calculated as,

$$-\frac{NI_c}{l_A} = \Delta H_A \Leftrightarrow I_c = -\frac{2H_c l_A}{N} \tag{5}$$

However, under poor contact conditions or increased air-gap length, the required magnetization current increases significantly. In extreme cases, such as when the magnetic flux path contains a large proportion of air, the required current may reach impractically high values for compact magnetization circuits. Moreover, once the excitation current ceases, the open-circuit flux of the NdFeB magnets may demagnetize the AlNiCo magnets if a low-reluctance magnetic flux path is not established, leading to magnetization failure.

These results indicate that reliable magnetization critically depends on both the pulse current and the contact condition. Therefore, precise regulation of the pulse current and contact-aware control is essential to ensure stable adhesion performance, which motivates the development of the magnetization control strategy described in the following section.

### *C. Performance comparison and optimization*

To fairly evaluate the performance of the proposed CHN-EPM, a comparative study was conducted with conventional S-EPM and P-EPM configurations, as shown in **Fig. 3**(a). The CHN-EPM adopts a circular Halbach-based topology, while S-EPM and P-EPM utilize square and parallel magnet arrangements, respectively. For a consistent comparison, all configurations were optimized under comparable outer dimensions and identical total permanent-magnet volume/mass. The allocation between NdFeB and AlNiCo was allowed to vary according to each topology during optimization.

TABLE I
THE OPTIMIZATION VARIABLES OF THE CHN-EPM

| Component | Parameter | Proposed CHN-EPM | S-EPM | P-EPM |
|---|---|---|---|---|
| NdFeB | Length *L* | 16 mm | 21 mm | 42 mm |
| | Width *W* | 8.2 mm | 8 mm | 8 mm |
| | Height *H* | 6.2 mm | 21 mm | 21 mm |
| | Mass | 37 g | 55 g | 55 g |
| AlNiCo | Length *L* | 32 mm (radius) | 21 mm | 42 mm |
| | Width *W* | 14.6 mm | 8 mm | 8 mm |
| | Height *H* | 8.2 mm | 21 mm | 21 mm |
| | Mass | 73 g | 55 g | 55 g |
| Yoke(DT4C) | Length *L* | 32 mm (radius) | 22 mm | 86 mm |
| | Width *W* | 16 mm | 22 mm | 18 mm |
| | Height *H* | 8.2 mm | 25 mm | 25 mm |
| | Mass | 98 g (Fan-yoke)<br>64 g (Disc-yoke) | 363 g | 580 g |
| Total | Total Mass | 272 g | 473 g | 690 g |

To optimize the geometric parameters of the three configurations, a simulation-driven approach based on finite element analysis (FEA) is adopted. Specifically, the electromagnetic performance is evaluated using JMAG 23, and a multi-objective genetic algorithm (MOGA) is employed to search for optimal design parameters. The optimization considers three main objectives: (1) maximizing the adhesion force in the ON state, (2) minimizing the residual force in the OFF state, and (3) minimizing the total mass. The optimized geometric parameters of the three configurations are summarized in **Table I** and are used consistently in subsequent simulations and experiments. As shown in **Table I**, under identical permanent magnet volume, S-EPM and P-EPM require significantly more yoke material to achieve optimal performance. In contrast, the proposed CHN-EPM attains comparable or superior performance with substantially reduced yoke usage, indicating a more efficient magnetic circuit design, and resulting in the lightest overall structure.

**Fig. 3** (b)–(d) illustrate the magnetic flux distributions of the three configurations under adhesion (ON) and de-adhesion (OFF) states. It can be observed that the CHN-EPM exhibits a more concentrated magnetic flux toward the contact interface, with significantly reduced flux leakage to the surrounding region. This advantage originates from its three-dimensional magnetic circuit topology. Compared with conventional S-EPM and P-EPM designs, which rely on planar flux paths and longer yoke conduction paths, the CHN-EPM forms a compact 3D flux loop with shortened magnetic paths. Meanwhile, its structural configuration provides a larger effective contact area with the ferromagnetic surface. In the OFF state, the CHN-EPM exhibits significantly reduced residual magnetic flux compared to S-EPM and P-EPM, indicating a clearer magnetic switching behavior between adhesion and de-adhesion states.

The quantitative comparison of adhesion force as a function of air-gap length is presented in **Fig. 3** (e). The CHN-EPM consistently achieves higher adhesion force density across the entire air-gap range. Moreover, the degradation rate of adhesion force with increasing air gap is significantly lower than that of S-EPM and P-EPM, demonstrating improved robustness to air-gap variations and contact uncertainty. This observation is consistent with the theoretical analysis in Section II-B, where the air-gap flux density of the CHN-EPM is less sensitive to air-gap variations.

In addition, from the improvement ratio of adhesion force density in **Fig. 3** (f), the CHN-EPM achieves a substantial improvement, reaching up to 335% higher than S-EPM and 450% higher than P-EPM. This significant enhancement can be attributed to two key factors: 1) the flux concentration effect enabled by the three-dimensional magnetic circuit, and 2) the increased effective contact area resulting from the structural arrangement. These features allow the CHN-EPM to achieve higher adhesion performance with reduced material usage and mass, implying a significantly higher load-to-weight ratio at the system level.

Overall, the results demonstrate that the proposed CHN-EPM significantly outperforms conventional EPM configurations in terms of adhesion force, force density, and robustness to air-gap variations, validating its effectiveness for high-load-density and reliable wall-climbing robotic applications.

## III. MAGNETIZATION CONTROL SYSTEM FOR MAGNETIC FOOT

To achieve reliable and controllable adhesion, a magnetization control system is developed to regulate the pulse current amplitude and duration under varying contact conditions. The hardware architecture and pulse current generation are presented, followed by the magnetization control strategy.

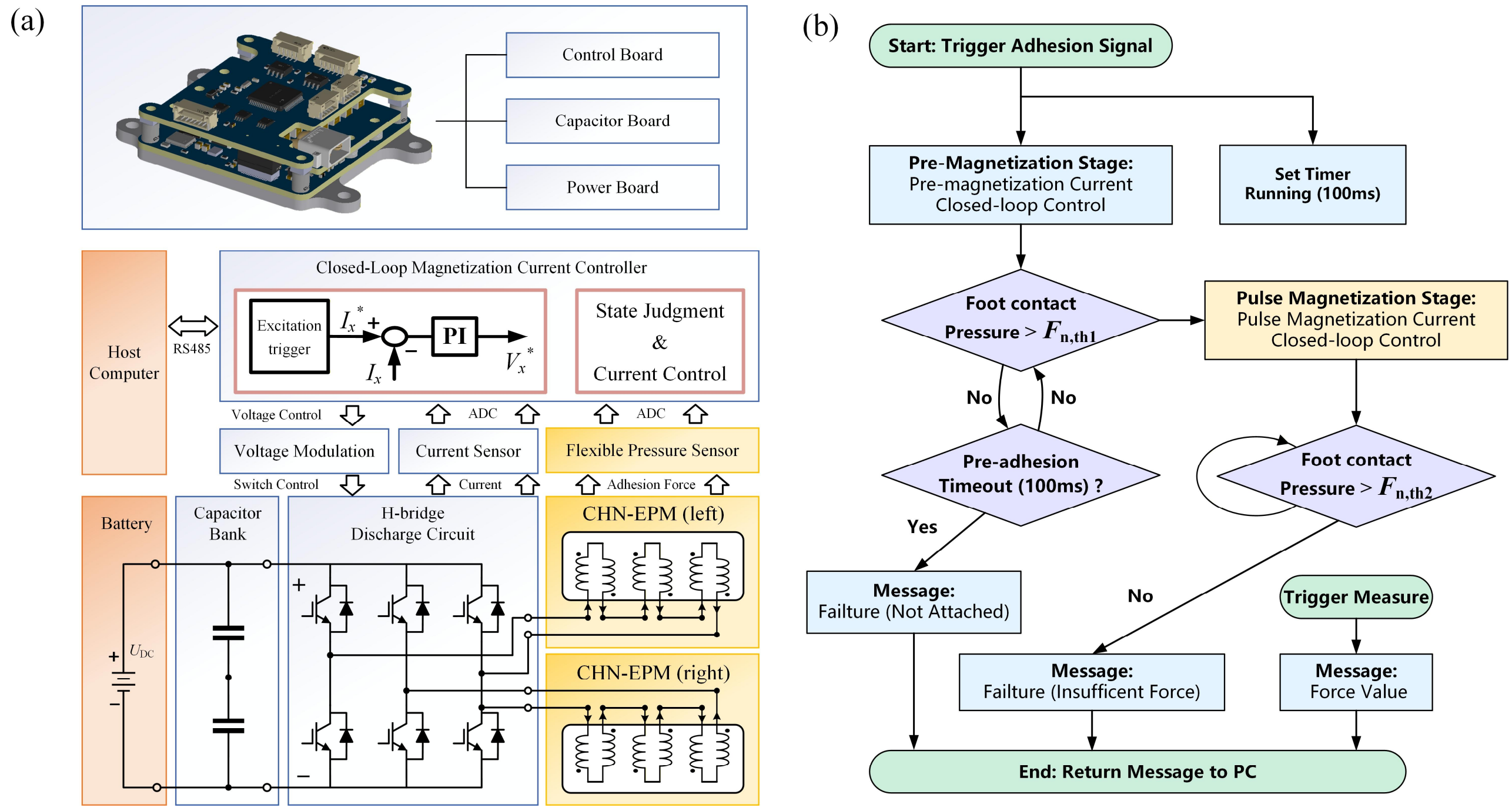

**Fig. 4.** Magnetization control system of the magnetic foot: (a) hardware architecture and pulse current generation module; (b) magnetization control strategy with force-feedback-based decision logic.

### *A. Hardware Architecture and Pulse Current Generation*

The hardware architecture of the magnetization control system is illustrated in **Fig. 4**(a). To achieve compact and efficient excitation of the CHN-EPM, a dedicated magnetization control module is developed, consisting of three stacked circuit boards: a control board, a capacitor board, and a power board. The hardware is designed for a 48 V DC input and can deliver pulse currents of up to 50 A. With a compact form factor (62 g), it enables seamless integration into quadruped robotic platforms.

The control board coordinates system operation and executes the magnetization control logic, including pulse current regulation, current monitoring, and communication with the upper-level controller. The capacitor board serves as the energy storage unit, providing short-duration high-power pulses required for magnetization. The power board implements a three-phase H-bridge circuit, which converts the stored energy into bidirectional pulse currents for magnetization and demagnetization of the CHN-EPM. During operation, once receiving a control command, the stored energy from DC power supply and capacitor bank is released through the H-bridge to generate a high-current pulse in the excitation coil. By modulating the leg voltage of the bridge, both the direction and duration of the pulse current can be regulated.

To improve system compactness and power density, a shared-bridge topology is proposed for dual-foot actuation, in which two CHN-EPM units share part of the H-bridge, allowing a single driver with three bridge arms to control two magnetic feet. Consequently, only two magnetization control modules are required to operate all four feet of the quadruped robot. To ensure independent operation of the two units under the shared hardware, a time-division modulation strategy is adopted. When one unit connected between arms 1 and 3 is excited, the corresponding bridge arms are actively modulated, while the remaining arm (arm 2) is driven synchronously with the shared arm (arm 3). This coordinated operation suppresses the voltage difference across the inactive unit (between arms 2 and 3), thereby preventing unintended excitation currents and ensuring decoupled and reliable operation.

### *B. Magnetization Control Strategy*

The magnetization control strategy is illustrated in **Fig. 4** (b). To ensure reliable magnetization under varying contact conditions, a two-stage control strategy combining current regulation and force-feedback evaluation is adopted.

Upon receiving a magnetization command, the system first enters a pre-magnetization stage, where a low-level current (approximately half of the nominal magnetization current) is applied continuously. Although full magnetization is not achieved at this stage, a basic adhesion force is generated, allowing the system to evaluate the contact condition. The contact force is monitored using the embedded pressure sensor, and once it exceeds a predefined threshold $F_{n,\,th1}$, sufficient contact with the ferromagnetic surface is confirmed.

The system then transitions to the pulse magnetization stage, where a high-current pulse (complete nominal magnetization current) is applied under closed-loop current control to fully magnetize the AlNiCo magnet. After excitation, the adhesion state is evaluated by comparing the measured contact force with a second threshold $F_{n,\,th2}$, which corresponds to the required adhesion level for safe operation. To enhance robustness, the system performs autonomous success evaluation at each stage. If the contact force does not reach $F_{n,\,th1}$ within a predefined time window (e.g., 100 ms), the process is terminated and identified as a contact failure. Similarly, after pulse magnetization is completed, the adhesion force is continuously monitored. If it falls below $F_{n,\ th2}$, an adhesion failure is detected, and the evaluation results are communicated to the upper-level controller for further decision-making. In addition, the upper-level controller can also directly access the real-time force measurements.

By integrating pre-magnetization, closed-loop pulse control, and multi-threshold force evaluation, the proposed strategy enables reliable and repeatable magnetization under uncertain contact conditions, thereby improving the safety and robustness of the magnetic foot during wall-climbing locomotion.

## IV. Experimental Verification

### *A. Experimental Setup and Methodology*

As shown in **Fig. 5** (a)–(b), a test setup was constructed to evaluate the CHN-EPM magnetic foot module. The module was mounted on a linear-guide carriage and brought into contact with a Q235 steel plate rigidly fixed at the right end of the fixture. The carriage was connected in series with a high-precision load cell based on strain-gauge technology (DYLY-07, 200 kg range, accuracy 0.03% full scale) and a linear actuator anchored to the left-end aluminum frame.

By adjusting the relative orientation between the magnetic foot and the steel plate, two test modes were enabled, namely normal pull-off for adhesion force measurement and tangential shear for friction force evaluation. In addition, Q235 steel plates with different thicknesses were prepared to investigate the influence of contact conditions on adhesion performance.

The magnetization driver was powered by an independent 48V lithium battery pack, rather than an external bench power supply. Although the driver hardware allows higher peak current, the closed-loop current commands used in the experiments were limited to 16 A for full magnetization and 12 A for demagnetization. The transient coil current and pulse width were measured using a HIOKI MR6000 waveform recorder with a current clamp, enabling accurate characterization of the magnetization and demagnetization dynamics.

### *B. Performance Evaluation of the Magnetic Foot*

**Fig. 5** (c) first presents the adhesion force as a function of pulse current amplitude under different wall thicknesses (2, 4, 6, and 8 mm). As the current amplitude increases, the adhesion force correspondingly rises and gradually reaches saturation beyond approximately 16 A, demonstrating that the adhesion level can be effectively regulated through current control. Such controllability enables adaptive operation under varying working conditions: lower current amplitudes can be applied under light-load scenarios to reduce energy consumption and increase switching frequency, while higher amplitudes can be used under heavy-load or long-duration conditions to ensure reliable adhesion performance. Correspondingly, the energy required for each magnetization/demagnetization process is approximately 15 J for maximum adhesion force, and can be reduced to around 5 J under lower-load scenarios by adjusting the pulse current amplitude and duration, highlighting the energy-efficient and adaptive characteristics of the proposed CHN-EPM system. **Fig. 5** (c) further illustrates the relationship between adhesion force and air gap under a fixed wall thickness of 10 mm and a pulse current amplitude of 16 A. As the air gap between the magnetic foot and the ferromagnetic surface increases, the adhesion force gradually decreases. Notably, the reduction exhibits an approximately linear trend, which is consistent with the simulation results, indicating reduced sensitivity to air-gap variations.

**Fig. 5** (d) illustrates the influence of wall thickness on the adhesion and friction performance of the CHN-EPM module. As the thickness of the Q235 steel plate increases, both the normal holding force and the tangential friction force rise rapidly at first and then gradually approach saturation beyond approximately 6 mm. The maximum adhesion force reaches about 1200 N, while the maximum friction force approaches 660 N. This trend can be attributed to the magnetic circuit characteristics. When the wall is relatively thin, magnetic flux tends to saturate within the ferromagnetic material, leading to increased magnetic reluctance and reduced effective flux, thereby limiting the adhesion force. As the wall thickness increases, the magnetic circuit becomes more complete and the flux distribution improves. Once the thickness exceeds a certain threshold, magnetic saturation is effectively avoided, and the adhesion force reaches its maximum and remains nearly constant. In addition, the friction force follows a similar trend to the normal adhesion force, as it is directly proportional to the magnetically induced normal contact force.

**Fig. 5** (e) presents the measured pulse current profiles during the magnetization and demagnetization processes, together with the corresponding foot-end force response. During magnetization, a two-stage pulse current strategy is adopted. The excitation current first rises to approximately $0.5I_{mag}$ (8 A). Once the measured foot-end force exceeds a predefined threshold $F_{\mathrm{n,th1}}$ (500 N), the current is rapidly increased to the maximum magnetization current $I_{mag}$ (16 A), driving the force to exceed 1000 N. As a result, the entire magnetization process is completed within 35 ms. In addition, a second threshold $F_{\mathrm{n,th2}}$ is set to 800 N for post-magnetization monitoring; as long as the adhesion force remains above this threshold, no failure is reported. This strategy ensures reliable adhesion while avoiding prolonged high-current excitation, thereby significantly reducing energy consumption. During demagnetization, a controlled current profile is applied, allowing the demagnetization current $I_{demag}$ (12 A) to reach its target within 20 ms. Based on prior experimental observations, this current level is sufficient to fully eliminate the residual magnetization, enabling a reduction in demagnetization current and further lowering energy consumption. These results demonstrate that precise current regulation is critical for achieving reliable magnetization and demagnetization while maintaining high energy efficiency, which strongly relies on the proposed magnetization driver rather than uncontrolled direct discharge.

It is worth mentioning that each measurement point in **Fig. 5**(c)–(d) was repeated five times. The plotted values represent the mean, and the error bars indicate the standard deviation. Overall, the results show that the proposed system enables controllable magnetization while maintaining strong and stable adhesion under varying conditions, thereby providing experimental validation for the proposed closed-loop magnetization strategy, and highlighting the importance of integrating pulse-current control with contact-state awareness.

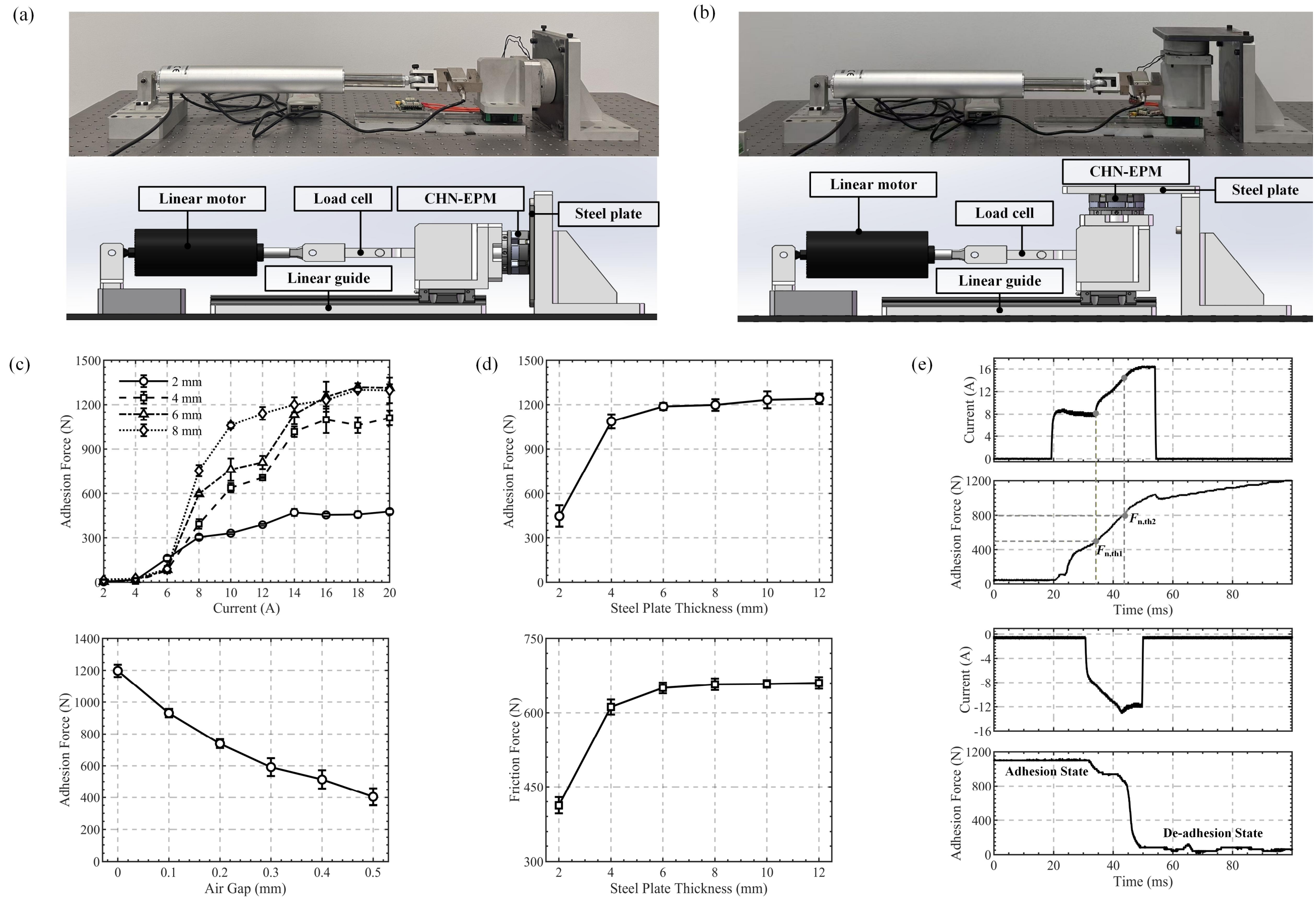

**Fig. 5.** Experimental setup and characterization of the CHN-EPM magnetic foot: (a) normal force test platform; (b) friction force test platform; (c) adhesion force vs pulse current amplitude and air gap under different wall thickness; (d) adhesion and friction forces vs wall thickness; (e) pulse current and force sensing profiles during magnetization/demagnetization.

### *C. System-Level Validation on the Quadruped Climbing Robot*

For system-level validation, four CHN-EPM modules and two magnetization drivers were integrated into a Unitree GO2 platform, as shown in **Fig. 1**. The integrated robot has an overall size of approximately 700 × 310 × 400 mm and a system weight of about 18 kg. The onboard controller is an NVIDIA Jetson Xavier NX, and the control framework runs at 500 Hz. The locomotion controller is implemented based on the MPC framework in [22], including the state estimator, magnetic foot controller, and balance controller. Since this work focuses on the magnetic adhesion foot rather than perception-based foothold planning, predefined MPC-based locomotion patterns were used in the experiments. As shown in **Fig. 6** (a)–(c), both single-foot and whole-body load-bearing tests were conducted. A single CHN-EPM magnetic foot was attached to a ceiling steel plate and used to support a 100-kg dumbbell load, consisting of four 10-kg plates and three 20-kg plates. For whole-body adhesion tests, the 18-kg robot stably supported an additional 20-kg load on the ceiling and an additional 8-kg load on the vertical wall. In comparison, the MARVEL robot reported in [22] has a self-weight of approximately 8 kg and demonstrated payloads of 3 kg on ceilings and 2 kg on vertical walls. These results show that the proposed CHN-EPM system provides a large adhesion margin for heavy-load quadruped climbing.

Vertical climbing experiments were further conducted on a painted steel wall, as shown in **Fig. 6** (d). The wall was made of 8-mm-thick Q235 steel and coated with primer, intermediate paint, and topcoat, with a total paint thickness of approximately 0.4 mm. This coating introduces an additional effective air gap between the magnetic foot and the steel substrate, similar to painted bridge-like ferromagnetic structures. The robot achieved stable vertical climbing at a speed of approximately 0.12 m/s, indicating reliable adhesion under relatively large air-gap conditions.

To evaluate performance under partial-contact conditions, the robot was tested on a vertical perforated Q235 steel plate, as shown in **Fig. 6** (e). The perforated plate has a thickness of 5 mm, a hole diameter of 10 mm, and a hole spacing of 20 mm. Since the outer radius of the magnetic foot is approximately 40 mm, each foot covers about nine holes during contact, reducing the effective contact area and creating discontinuous magnetic flux paths. Despite this condition, the robot maintained stable climbing locomotion, demonstrating the robustness of the distributed CHN-EPM topology and force-feedback-based magnetization strategy under partial contact.

Finally, curved-surface locomotion was tested on a vertical semicylindrical ferromagnetic surface with a radius of 2 m, as shown in **Fig. 6** (f). The steel plate thickness is 3 mm, and the paint coating thickness is approximately 0.1 mm. Stable locomotion on this curved and relatively thin steel surface verifies the surface adaptability of the passive ankle mechanism and the ability of the magnetic foot to operate under reduced wall thickness and non-planar contact conditions.

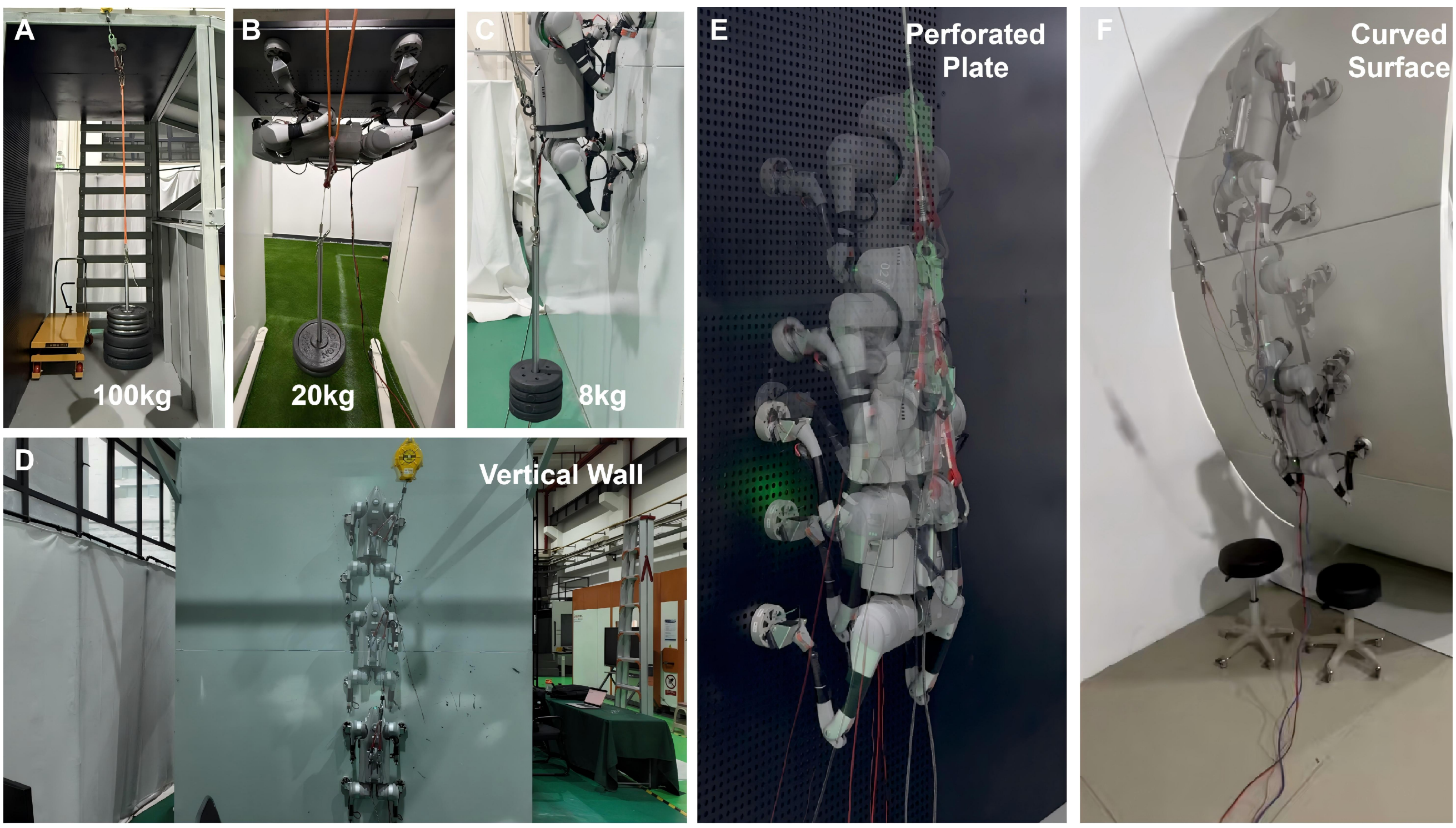

**Fig. 6.** System-level validation of the quadruped wall-climbing robot with the proposed CHN-EPM magnetic feet. The corresponding demonstrations are provided in Movies S0–S3. (a) Single-foot adhesion test with a 100-kg load. (b) Whole-body adhesion test on a ceiling with a 20-kg external load. (c) Whole-body adhesion test on a vertical wall with an 8-kg external load. (d) Vertical climbing on a painted steel wall. (e) Vertical climbing on a perforated steel plate under partial-contact conditions. (f) Vertical climbing on a curved ferromagnetic surface.

Overall, the system-level experiments show that the proposed CHN-EPM magnetic foot enables high-load static adhesion and stable locomotion on painted, perforated, and curved ferromagnetic surfaces. The corresponding videos are provided as supplementary materials: Movie S0 for the load-bearing tests, Movie S1 for vertical climbing on the painted flat wall, Movie S2 for climbing on the perforated plate, and Movie S3 for climbing on the curved surface. Additional implementation materials, selected 3D structural models, demonstration videos, and related documentation are publicly available in [30].

### D. Discussion and Comparison

**Table II** compares representative magnetic foot mechanisms in terms of energy consumption, adhesion regulation, switching speed, contact awareness, and tolerance to air gaps. PM-based feet provide strong and energy-efficient attachment, but their adhesion is not actively controllable and detachment usually requires additional mechanical actuation. EM-based feet allow fast and fully controllable switching, but they rely on continuous power supply and thus suffer from high energy consumption. Conventional EPM-based feet achieve low-power switching, yet their adhesion is generally limited to binary on/off states and is typically driven by open-loop pulse excitation, making them sensitive to contact uncertainty and air-gap variations.

By contrast, the proposed CHN-EPM system enables not only switchable but also tunable adhesion through closed-loop pulse current control while preserving the low-energy advantage of EPMs. Moreover, by integrating foot-end force sensing into the magnetization loop, the proposed system introduces contact-aware adhesion control, which improves switching reliability under imperfect contact conditions. As a result, it achieves a more favorable balance among low power consumption, fast switching, controllable adhesion, and robustness to air gaps.

## V. CONCLUSION

This paper presents a novel magnetic foot for climbing quadruped robots, integrating a force-feedback CHN-EPM with a closed-loop magnetization control strategy. The proposed CHN-EPM enhances adhesion performance through a parallel magnetic flux network, while the introduced force-feedback-based control enables precise regulation of the magnetization process under uncertain contact conditions. Experimental results at both module and system levels demonstrate that the proposed method achieves strong adhesion, fast and reliable magnetization switching, and robustness to air gaps and partial contact. Stable locomotion was validated on painted vertical walls, perforated plates, and curved ferromagnetic surfaces, while high-load adhesion was demonstrated on ceiling surfaces, confirming the effectiveness of the approach in practical scenarios. Overall, this work addresses the coupled challenge of magnetization controllability and contact uncertainty in compact magnetic feet, providing a reliable and energy-efficient solution for climbing robots in industrial environments. Future work will focus on further weight reduction, energy optimization, and adaptive control strategies for more complex and unstructured surfaces.

TABLE II COMPARISON OF DIFFERENT MAGNETIC ADHESION MECHANISMS AND CONTROL CAPABILITIES

| Magnetic Foot Type | Energy Consumption | Adhesion Regulation | Switching Speed | Contact Awareness | Air-Gap Tolerance |
|---|---|---|---|---|---|
| Permanent Magnet (PM) in [17] | Medium (detachment requires power) | Not controllable | Slow | ✗ No | ✓ High |
| Electromagnet (EM) in [13] | High (Continuous power) | Fully controllable | Fast | ✗ Limited | ✓ Moderate |
| Electro-Permanent Magnet (EPM) in [22] | Low (switching only) | Switchable | Medium | ✗ No | ✗ Low |
| Electro-Permanent Magnet (EPM) in [23] | Low (switching only) | Switchable | Medium | ✗ No | ✗ Low |
| Proposed CHN-EPM | Low (switching only) | Switchable & tunable | Fast | ✓ Integrated | ✓ Moderate |